# Towards the Ultimate Programming Language: Trust and Benevolence in the Age of Artificial Intelligence

Bartosz Sawicki, Michał Śmiałek, Bartłomiej Skowron[1]

Warsaw University of Technology
July 2024

**Abstract:**
This article explores the evolving role of programming languages in the context of artificial intelligence. It highlights the need for programming languages to ensure human understanding while eliminating unnecessary implementation details and suggests that future programs should be designed to recognize and actively support user interests. The vision includes a three-level process: using natural language for requirements, translating it into a precise system definition language, and finally optimizing the code for performance. The concept of an "Ultimate Programming Language" is introduced, emphasizing its role in maintaining human control over machines. Trust, reliability, and benevolence are identified as key elements that will enhance cooperation between humans and AI systems.

## How the Sofware is Developed?

A computer program is classically presented to its users as a mysterious box that processes input data to produce the expected output. This simplification is a good starting point as it highlights the crucial role of input data and the fact that the processing mechanism is not important as long as the output meets expectations. However, it is also necessary to describe the methods of communication with the user or other cooperating systems, as well as to build an internal data model. This is the basis for the dominant multi-layered architectures today, such as the Model-View-Controller or Model-View-Presenter patterns.

Software is created by people and ultimately for people [2]. They use programming languages to describe the desired way of processing data. However, over the past 70 years, the way programs are defined has changed dramatically. Programming languages have undergone a significant transformation from machine code and assembly language to modern third generation (3G) languages, which offer high-level abstraction to facilitate the description of complex systems.

It should also be emphasized that a program is the final product of the entire software engineering process, which begins with the formulation of often complex requirements and the design of basic system components. Requirements are traditionally formulated in natural language, although various formal methods (controlled language, graphic models) are sometimes used. Requirements describe both the expected behavior of the system and the data structures processed by the system. In simplified terms, the entire software development process can be seen as transforming requirements into working machine code (a program) that meets these requirements.

## How Does AI Create Code?

Today, generative artificial intelligence systems can create software based on short descriptions formulated in natural human language. They rely on vast repositories of source code and other documented knowledge, which have been used to train an artificial neural network containing trillions of parameters [1].

However, the challenge lies in the fact that descriptions in natural language are ambiguous, just like the language used by humans. At the same time, models built on statistical foundations do not guarantee deterministic behavior. Therefore, there is no guarantee that the result will always meet expectations.

AI methods can generate a complete computer program based on a fairly general query in natural language. This means that all algorithmic components, the internal data model, and the user interface view are generated [5]. However, it seems that this approach is not optimal. It should be noted that a typical requirements specification includes dozens of pages of descriptions of functional units, domain concepts (glossary), domain models, quality requirements, and other necessary system features. The question arises as to how precise and unambiguous such a specification must be for the AI system to reliably produce a program that meets user expectations.

---

[1] bartosz.sawicki@pw.edu.pl, michal.smialek@pw.edu.pl, bartlomiej.skowron@pw.edu.pl

## What is the Role of Programming Language?

In the compilation process, a program's source code is translated into machine code tailored to the processor architecture on which it is executed. Simultaneously, it serves as a crucial communication tool between programmers. The ability to understand, control, and correct the way a program operates builds trust in the computer program and facilitates collaboration.

A longstanding problem is that the system specifications created by users are highly variable. People often cannot predict what outcome will be best, and external circumstances change as well. This necessitates that the software development process allows for modifications and improvements to the source code. Iterative software development methodologies like Scrum respond to this need, enabling users to regularly influence the appearance and functionality of the software being developed [7].

The reliability of today's processing methods largely stems from established data models on which they are based. A consistent data model allows several independent applications to operate on them and mutually control the correctness of their operations.

This raises the question of what programming will look like in the era of strong artificial intelligence. It seems we are approaching the ultimate generation of programming languages, where the source code will encapsulate the essence of user expectations for the system [8].

## Trust, Reliability, and Benevolence of a Software

Let us assume for a moment that a software user is a citizen of a certain country, and the software is a set of regulations (laws) according to which public institutions of that country operate. When does a citizen trust a public institution? When the institution is reliable. If it is not reliable, the citizen questions its competence, the safety of the area under its jurisdiction, and doubts the public good that the institution should uphold. When a citizen does not trust, they are not only critical of the institution but also do not support its operations and sometimes even oppose it. When they trust, the chance for mutual cooperation increases. Social studies indicate that the reliability of public institutions is built on factors such as benevolence, competence, and integrity of state institutions [6]. The counterparts of integrity and competence in computer programs are the aforementioned abilities to understand, control, and correct the way a program operates. Nevertheless, extending this analogy further, can software be benevolent?

Aristotle defined benevolence from one person to another not as politeness, being nice, smiling, and cordiality, as we spontaneously tend to think, but as understanding what is good (what contributes to the flourishing and good life of that person) and striving to achieve that person's good. A benevolent person wishes well for another person and acts for the good of the other person. Mutual benevolence is a necessary condition for friendship. To be benevolent, I must understand the other party's interest and actively want to help realize that interest—mere observation is not enough. A benevolent state institution is one that sees the citizen's interest and is ready to actively work for the citizen's interest.

The potential evaluation of a program's benevolence, using this analogy, is the assessment of how well the software can understand the user's and the social group's good, of which the user is a part, and work to achieve that good. In other words, a benevolent AI is one that possesses a semblance of good will. Notice that when a certain person always meets our needs, meaning their behavior fully corresponds to our needs, it does not necessarily mean that we trust that person. On the contrary, we often become suspicious of them. We do not become suspicious of all people who make mistakes that work against us, but of those who make such mistakes and wish us ill. If someone wishes us well, we will more easily forgive them for a mistake, even if that mistake is severe for us. This is the trust-building power of benevolence. Similarly, with a computer program: if it makes mistakes, we still trust its operation if it is benevolent—because if it wishes us well, it will sooner or later actively correct its error. Therefore, the benevolence of AI is something that can enhance trust in even an imperfect AI.

## Vision of the Ultimate Programming Language

The need for trust is a critical element of cooperation between humans, and therefore also between humans and artificial intelligence systems. It seems that a programming language must provide humans with the ability to understand the method of operation while eliminating all unnecessary implementation details. Moreover, we propose that the program should, as far as possible, be benevolent to potential users by adequately recognizing their interests and actively supporting them.

Imagining further with the analogy to state structures, we envision that future source code will be an essential description akin to today's legal language, which will describe methods of data processing. Additionally, there will be a description of key internal data structures, enabling external audits of data integrity. The user interface is an artifact easily verifiable by humans, so it seems possible to give the AI model greater freedom in this area.

Three fundamental issues can be distinguished here: 1) how to formulate requirements so that AI understands the needs well and can be benevolent to the user; 2) what the programming language should look like to allow humans to easily modify what the AI has generated; 3) what language should be compiled and executed by computers. This raises the question of whether we will still need 3G programming languages.

Our vision includes three levels at which the AI-assisted software development process would take place. At the first level, a natural language with specified guidelines would be used. It would serve to define the problem in a way convenient for humans (functional requirements, domain concept glossary, data descriptions, quality requirements, definition of the well-understood interests of the user group). This language could be supported by various patterns, templates, or graphical notations, as well as ethical elements. Next, at the second level, strong AI language models would translate this description into a new system definition language. This language would be precise, unambiguous, and complete. It would be fully understandable to humans, yet also capable of being automatically and deterministically transformed into a traditional program in a 3G language.

At the second level, a dialogue between AI and humans would be possible - modification of the specification and adaptation to changing needs. This can be seen as a significant extension to the model-driven and low-code software development paradigms [7,9]. The translation from the second to the third level should be deterministic. The third level would enable code optimization in situations where, for example, system performance requires it.

The language placed at the second level of the above description we can call the Ultimate Programming Language. If humans want to maintain control over the operation of machines, further reduction of complexity seems impossible.